\documentclass[12pt]{iopart}

\usepackage{algorithm}
\usepackage{algpseudocode}
\usepackage{graphicx} 
\usepackage{subcaption}
\usepackage{url}
\usepackage{booktabs}
\usepackage{caption}
\expandafter\let\csname equation*\endcsname\relax
\expandafter\let\csname endequation*\endcsname\relax
\usepackage{amsmath}
\usepackage{array}
\usepackage{siunitx}
\usepackage[utf8]{inputenc}	
\usepackage{amsthm,amsfonts,amssymb,amscd}
\usepackage{multirow,booktabs}
\usepackage[table]{xcolor}
\usepackage{fullpage}
\usepackage{lastpage}
\usepackage{enumitem}
\usepackage{fancyhdr}
\usepackage{mathrsfs}
\usepackage{wrapfig}
\usepackage{setspace}
\usepackage{calc}
\usepackage{multicol}
\usepackage{cancel}
\usepackage[retainorgcmds]{IEEEtrantools}
\usepackage[margin=3cm]{geometry}
\usepackage{empheq}
\usepackage{framed}
\usepackage[most]{tcolorbox}
\usepackage{xcolor}
\usepackage{hyperref}

\newcommand{\R}{\mathbb{R}}

\newcommand{\set}[1]{\left\{#1\right\}}
\newcommand{\bb}[1]{\left(#1\right)}

\newcommand{\nll}{\text{NLL}}

\begin{document}
\UseRawInputEncoding
\title[Generative Pre-trained Transformer for Photoplethysmography Signals]{GPT-PPG: A GPT-based Foundation Model for Photoplethysmography Signals}

\author{Zhaoliang Chen$^1$, Cheng Ding$^2$, Saurabh Kataria$^3$, Runze Yan$^3$, Minxiao Wang$^3$, Randall Lee$^4$, and Xiao Hu$^3$}
\address{$^1$ Department of Computer Science, Emory University, Atlanta, GA, USA}
\address{$^2$ Department of Biomedical Engineering, Georgia Institute of Technology, Atlanta, GA, USA}
\address{$^3$ Nell Hodgson Woodruff School of Nursing, Emory University, Atlanta, GA, USA}
\address{$^4$ School of Medicine, University of California San Francisco, San Francisco, CA, USA}
\ead{david.chen2@emory.edu, xiao.hu@emory.edu}

\vspace{10pt}
\begin{indented}
\item[]February 2025
\end{indented}

\begin{abstract}
This study introduces a novel application of a Generative Pre-trained Transformer (GPT) model tailored for photoplethysmography (PPG) signals, serving as a foundation model for various downstream tasks. Adapting the standard GPT architecture to suit the continuous characteristics of PPG signals, our approach demonstrates promising results. Our models are pre-trained on our extensive dataset that contains more than 200 million 30s PPG samples. We explored different supervised fine-tuning techniques to adapt our model to downstream tasks, resulting in performance comparable to or surpassing current state-of-the-art (SOTA) methods in tasks like atrial fibrillation detection. A standout feature of our GPT model is its inherent capability to perform generative tasks such as signal denoising effectively, without the need for further fine-tuning. This success is attributed to the generative nature of the GPT framework. 
\end{abstract}

%
\vspace{2pc}

\section{Introduction}

The emergence of large language models (LLMs) such as BERT  \cite{DevlinJ} and GPT  \cite{RadfordA} has revolutionized the field of artificial intelligence by introducing the concept of foundation models. These models, characterized by extensive pre-training on large datasets without explicit supervision, demonstrate remarkable versatility across downstream tasks via fine-tuning. Foundation models are particularly transformative in clinical data analysis, where the availability of labeled training data is often limited. By leveraging their inherent flexibility and generalizability, foundation models offer a promising solution to these challenges  \cite{Hu2024}.

In the realm of time series data modeling, transformers have enabled significant advancements due to their ability to capture long-range dependencies and complex temporal patterns. Models like Informer  \cite{zhou2021informer} and Autoformer \cite{wu2021autoformer} have leveraged attention mechanisms to surpass traditional RNN-based approaches in forecasting accuracy. Meanwhile, generative transformer models such as GPT  \cite{RadfordA} have inspired general-purpose solutions like TEMPO  \cite{cao2024tempo} and TimeGPT  \cite{garza2023timegpt}, which employ pre-training and zero-shot learning techniques to adapt to diverse time series tasks. More recently, models such as PatchTST  \cite{NieY} and MOMENT  \cite{goswami2024moment} have focused on feature extraction and pre-training for time-series datasets. Additionally, EarthPT  \cite{SmithM} has applied GPT-like architectures to Earth Observation data, demonstrating the utility of transformers across various sequential domains.

The development of foundation models for physiological signals builds upon these advances, offering generalizable and transferable approaches for biosignal analysis. Models like HeartBeit  \cite{heartbeit} have utilized vision-based transformers for ECG analysis, while the Biosignal Transformer Model (BIOT)  \cite{biot} focuses on learning embeddings for EEG signals, facilitating cross-dataset transfer. Further innovations, such as the integration of latent representations from frozen LLMs with ECG reports  \cite{frozen_ecg}, and multimodal approaches such as SiamAF  \cite{guo2024siamaflearningsharedinformation}, highlight the adaptability of foundation models to novel biosignal datasets. Apple's large-scale wearable biosignal models  \cite{abbaspourazad2024largescaletrainingfoundationmodels} further underscore the transformative potential of foundation models in combining modalities like PPG and ECG for predictive tasks.

Within this broader context, photoplethysmography (PPG) emerges as an ideal modality for exploring foundation models. PPG signals are non-invasive, widely accessible, and easily collected through wearables and medical devices, making them suitable for foundational exploration  \cite{charlton2023wearable}. Deep learning methods have already achieved notable success in single-task applications such as atrial fibrillation detection  \cite{ding2024photoplethysmography}, heart rate estimation  \cite{IsmailS, jain2023selfsupervisedalgorithmdenoisingphotoplethysmography, 10385871}, and blood pressure prediction  \cite{ChowdhuryM}. Techniques like convolutional recurrent regressors  \cite{IsmailS}, cluster membership consistency methods  \cite{CMCPaper}, GAN-based augmentation for class imbalance  \cite{huang2023logspectralgan}, and signal quality-aware approaches like SQUWA  \cite{yan2024squwasignalqualityaware} and  \cite{Chen_2024} have further advanced the state-of-the-art in PPG signal analysis. Despite this progress, significant challenges remain in handling noise, motion artifacts, and dataset variability.

Recently, there has been a rise in developing foundation models for PPG. SiamQuality uses a convolutional neural network with contrastive learning techniques  \cite{Ding2024SiamQuality}. Meanwhile, PaPaGei employs both a patient contrastive approach and a morphology-aware self-supervised approach  \cite{pillai2024papageiopenfoundationmodels}. Both have demonstrated strong downstream capabilities. In this work, we present an extension of a preliminary version of GPT-PPG  \cite{chen2024gptppg}, a generative foundation model for PPG signals based on the GPT architecture. We demonstrate that our method not only excels at feature extraction to estimate physiological functions but can also perform signal denoising tasks. Specifically, our contributions are as follows:
\begin{enumerate}
\item We propose an adaptation of the GPT architecture to continuous PPG data and trained GPT-PPG at four different scales: 19M, 85M, 345M, and 1B parameters.
\item We propose a mixed-objective fine-tuning framework for downstream dataset adaptation, as well as extensions of this approach, including a fallback method for parameter-efficient fine-tuning and a bidirectional feature extraction method to allow for more robust performance.
\item We evaluated the model performance on a wide range of benchmarks, including atrial fibrillation detection, heart rate estimation, respiration rate estimation, and blood pressure estimation. We also provide qualitative evidence of how our model performs on signal denoising tasks.
\end{enumerate}

\section{Method}

\subsection{Data Preparation}

The dataset utilized for model pre-training was sourced from routine patient monitoring systems in the intensive care units at the University of California, San Francisco (UCSF) Medical Center. Data collection was conducted under an approved waiver of written patient consent, as sanctioned by the UCSF Institutional Review Board (IRB number: 14-13262). The analysis of this fully de-identified data complies with a data use agreement established between UCSF and Emory University.

This waveform dataset, encompassing physiological signals, cardiac arrhythmia alarms, and integrated electronic health records (EHR) from more than 24,100 patients, was gathered at UCSF Medical Center over the period of March 2013 to December 2018. The dataset comprises 2.6 million hours of continuous signal recordings, which include seven-lead ECG signals, a one-channel PPG signal, and a one-channel respiration rate signal. For the purposes of this study, we only use the one-channel PPG. 

To preprocess this data, we first segment the whole dataset into non-overlapping 30s strips and resample to 40Hz. We then normalized each sample $X \in \mathbb{R}^{1200}$ into $[0, 1]^{1200}$ by min-max normalization. That is
$$X = \frac{X-\min(X)}{\max(X)-\min(X)}$$
We use min-max normalization mainly because of 2 reasons: 1) it makes sure that the input range is fully bounded and 2) it does not make an underlying assumption about the distribution of the data points. However, we do note that other normalization schemes (such as z-score standardization) are possible and may change the training dynamics. 

Finally, reshape the normalized signal into 30 consecutive, non-overlapping patches, $X = \set{x_1, x_2, ..., x_{30}}$, with each patch $x_i \in \R^{40}$ encapsulating 1 second of the signal.

\subsection{Model Architecture and pre-training}

In leveraging the Generative Pre-trained Transformer (GPT) framework for analyzing continuous time-series PPG signals, we preserved the core components of GPT while making key adjustments to suit the unique characteristics of physiological signals. The decision to employ GPT was driven by its proven capability in capturing complex dependencies within sequential data, making it a promising candidate for time-series analysis. In addition, unlike models that are trained with mask-reconstruction objective such as BERT, GPT is more sample-efficient because in each iteration, the loss is computed for all token predictions rather than a random subset of them  \cite{clark2020electrapre-trainingtextencoders}\footnote{Normally, BERT is trained with reconstructing roughly 15\% of masked tokens, while in GPT, the loss is calculated for all tokens. }. We trained 4 models, GPT-19M, GPT-85M, GPT-345M, and GPT-1B. In the results section, we plot a scaling curve to uncover how model performance as we increase the model size. 

\subsubsection{Model Architecture}

Our model uses a simple yet effective linear layer to map each patch to a vector of dimension $d$. In the standard GPT implementation, we typically shift the entire sequence to the right by injecting a special token such as [SOS] in front to denote start of sentence and dropping the last token. We employ a similar technique. However, instead of injecting special numbers in front of the PPG sequence, a learnable vector $h_s$ of dimension $d$ is registered to the model directly and prepended to the embedded PPG sequence. Therefore, the input to the transformer backbone is $\set{h_s, h_1, h_2, ..., h_{29}}$. Note that we do not include $h_{30}$ in the model input since we do not have a ground truth $x_{31}$ to compare the model's output at this position with. This approach not only facilitates the model's training by providing a consistent starting point for each sequence but also avoids the appearance of an out-of-distribution patch (i.e., the [SOS] patch) that could otherwise impair the linear embedding layer's effectiveness. Since the signal is always 30s long in the pre-training phase, the traditional [EOS] (end of sequence) token is rendered unnecessary. 

Our stacked transformer decoder layer follows the common design choices of the popular GPT framework. Prior to attention module and feed-forward network (FFN), we employ root mean square normalization  \cite{ZhangB}. In addition, we use rotary positional embedding  \cite{SuJ}, to capture relative positional information of the sequence. 

\subsubsection{Prediction Head and Loss Function at Pre-training} In the realm of NLP, the GPT model has a classification head which predicts a token probability distribution. However, since PPG signals are continuous, we need to use a different prediction head and loss function. 

A natural choice is to predict the input patches directly, mapping hidden representations back to the input space and using a distance loss function like Mean Squared Error (MSE). However, our initial experiments with MSE led to model collapse, where the model predicts the mean value of the normalized signals regardless of the input sequence, which is a non-informative constant around 0.5. We conjecture that there are two main causes of this phenomenon. First, MSE has support over the entire real line while PPG data in our dataset is normalized to $[0,1]$ interval, which poses a significant distribution mismatch. Second, good quality PPG signals after normalization are typically rather symmetric about the mean, making predicting 0.5 an easy local minimum to converge to. 

To address this issue, we use a distribution-based loss function that has support over open interval $(0, 1)$, called logit-Laplace distribution loss  \cite{RameshA}. A Laplace distribution is parameterized by location parameter $\mu$ and scale parameter $b$. For each data point $x_i \in \R$, the model predicts real numbers $\mu_i, b_i$, and the objective is maximize the likelihood of observing $x_i$ given $\mu_i$ and $b_i$. The logit-Laplace distribution is given by applying a sigmoid transformation over a Laplace distributed random variable. Its probability density function (PDF) is given by:
\begin{equation}
    f(x; \mu, b) = \frac{1}{2bx(1-x)}\exp\bb{-\frac{|\text{logit}(x)-\mu|}{b}}
\end{equation}
where $x$ is ground truth, $\mu$ is the predicted location parameter and $b$ is the predicted scale parameter. It can be shown that minimizing the negative log likelihood of a Laplace distribution is equivalent to minimizing L1 distance loss (for more details, see \ref{A1}), and applying the sigmoid transformation restricts the support to $(0, 1)$ open interval, making the negative log of the above PDF a suitable loss function. Note that because the prediction is made for every data point, for a sequence of length $L$, the model prediction has dimension $2\times L$. In implementation, we simply use a linear layer to map each patch-level feature vector $h \in \R^D$ into output $o \in \R^{2P}$, where $P$ is patch size, which we then reshape and split into $\mu \in \R^P$ and $b \in \R^P$. We call this linear layer \textit{signal modeling head}. 

However, the normalization of signals to the $[0, 1]$ interval poses a challenge at the boundaries. Due to the nature of the min-max normalization, we are guaranteed that 0 and 1 will occur in each 30s sample, and either will cause the denominator $2bx(1-x)$ go to 0. To avoid this numerical issue, we perform an invertible transformation on $x$ to ensure that it is in $[0.1, 0.9]$ interval. This approach is favored over adding a small constant $\epsilon$ to the denominator $2bx(1-x)$, which will likely cause instability in training because then the term $1/\epsilon$ will occur in the loss function for every training sample. For a more detailed discussion about the intuition and assumptions of the training objective, please refer to \ref{loss_fn description}. 

\subsection{Mixed Objective Fine-tuning Framework}\label{mixed_objective}

\begin{figure}
    \centering
    \includegraphics[width=1.0\linewidth]{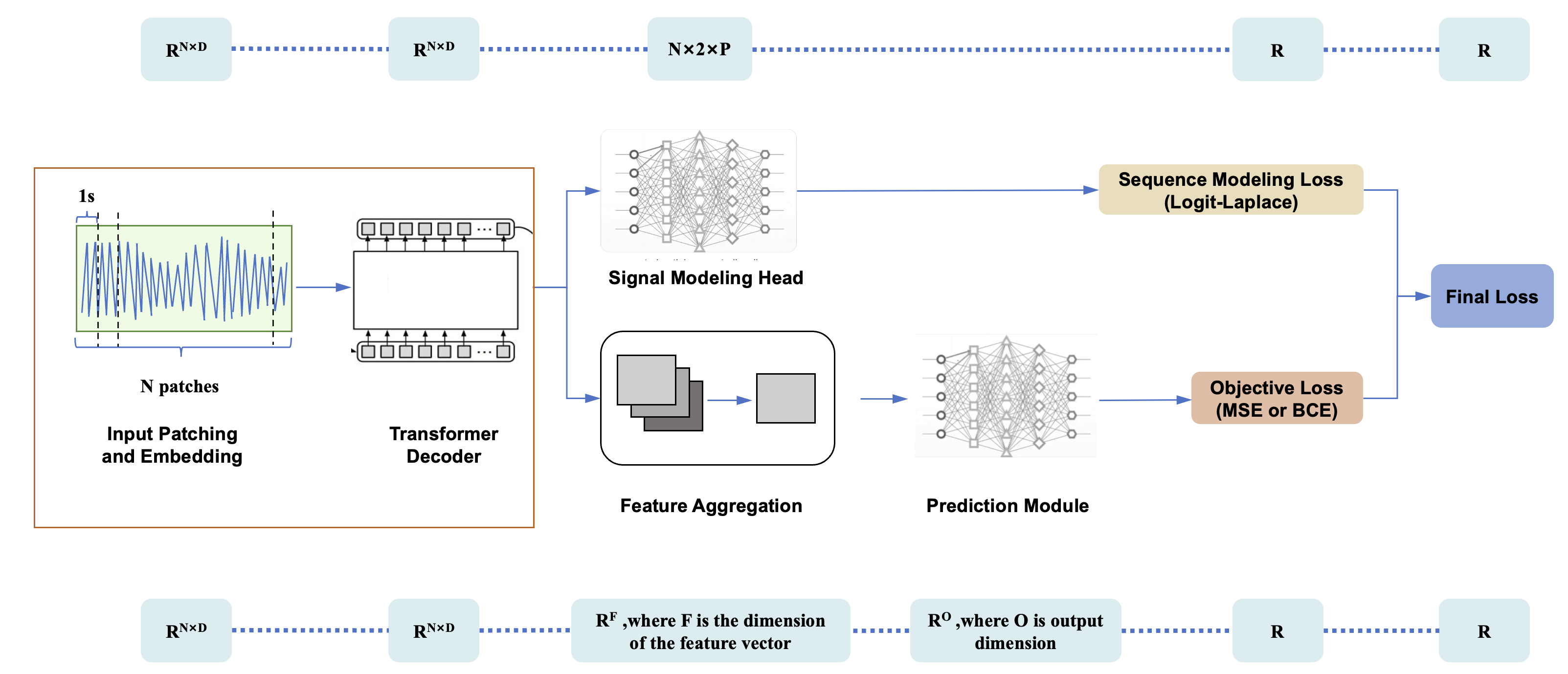}
    \caption{Mixed-objective Fine-tuning Framework}
    \label{f1}
\end{figure}

When fine-tuning the pre-trained GPT model for downstream tasks, we use a mixed-objective framework, where the loss function is a linear combination of the objective loss $L_o$, corresponding to the fine-tuning objective, and signal modeling loss $L_m$, which is logit-Laplace distribution loss obtained on the downstream dataset. Empirically, we found that when performing full-parameter fine-tuning, using the combined loss function is significantly better than using objective loss alone. Figure \ref{f1} provides a visualization of the framework, where $N$ is the number of patches, $P$ is the patch size, and $D$ is the hidden dimension of the model. The shape labeling beside each block refers to the output shape.

A key advantage of the GPT architecture is its adaptability to varying sequence lengths. This attribute is particularly beneficial in fine-tuning, where the sequence length may be different from the 30-second segments used in pre-training, allowing for seamless adjustment to the specific requirements of each task. However, due to the nature of a GPT network, we would have the same number of feature vectors as the number of patches. Therefore, prior to making a final prediction, it is essential that we first aggregate the features, shown on the lower branch of Figure \ref{f1}. We first explore a simple attention pooling mechanism as the feature aggregation method, and a multi-layer perception (MLP) as the prediction module. We discuss some extensions to this framework in Section \ref{extension}. 

\subsubsection{Attention-based Prediction Head} Given patch representations $H\in \R^{N\times D}$ generated by transformer decoder, the attention pooling layer $\text{AP}(\cdot)$ is expressed as:
$$\text{AP}(H) = \text{Softmax}(Hw)^\top H$$
where $w\in \R^{D\times 1}$ is a learnable parameter mapping each feature vector $H_i$ to a scalar. Essentially, $\text{AP}(\cdot)$ is a weighted sum of patch-level feature vectors, with weights normalized and learned by a neural network. This is a commonly used technique for feature aggregation. The resulting sequence level feature $h \in \R^{D}$ is then passed through a Gated Multi-layer Perceptron (Gated MLP), which is expressed as:
$$\text{GatedMLP}(h) = \text{SiLU}(h^\top W_2\odot h^\top W_3) W_1$$
where $W_1 \in \R^{D'\times O}, W_2 \in \R^{D \times D'}$, $W_2 \in \R^{D \times D'}$ are learnable parameters, SiLU refers to the Sigmoid Linear Unit activation function, $O$ refers to output dimension, and $D'$ is the hidden dimension of MLP. Gated MLPs are frequently used in modern language models due to its expressiveness. In this work, since our downstream tasks are either regression or binary classification, the output dimension $O$ is set to be 1 and 2, respectively. 

\subsubsection{fine-tuning Process} In our fine-tuning framework, we synergize the objective loss $L_o$ with the signal modeling loss $L_m$ to formulate a combined loss function: $L(y,y',X,X')=L_o(y,y')+\lambda L_m (X,X')$, where $X,y$ are PPG signals and labels of the dataset and  $X',y'$ are model predictions of the signal and label. In code implementation, we first get patch representations $H \in \R^{N \times D}$ from the GPT decoder. We feed it through both signal modeling head and attention-based prediction head to get outputs necessary for computing the above loss, as Figure \ref{f1} demonstrates. Empirically, we found that slowly annealing $\lambda$ to 0 was more beneficial than fixing it throughout the fine-tuning process. 

We observed that this delivered much better performance and faster convergence compared to using objective loss only. The auxiliary signal modeling loss helps GPT-PPG to learn the distribution of the PPG signals in the downstream dataset, which could be very different from the distribution of the pre-training dataset. This improves the representation generated by the GPT-PPG model, which in turn facilitates better predictions. 

For regression tasks, we use the standard MSE as objective loss. Note that logit-Laplace loss does not apply here because the labels of the regression task (such as heart rate estimation) are not normalized to $[0, 1]$. For classification tasks, we use cross entropy as per standard. Signal modeling is done in the same way as it is in the pre-training phase, and hence uses logit-Laplace distribution loss. 

\subsection{Extensions of the Fine-Tuning Framework}\label{extension}

The framework proposed in Figure \ref{f1} can be very flexible. Indeed, based on specific needs, we may use different feature aggregation methods, prediction modules, and even manipulating the behavior of transformer decoder. The key idea behind the mixed-objective framework is to take advantage of both signal modeling loss, which can be interpreted as the likelihood of the given data sample falling into the pre-training distribution, and the objective loss. There are naturally different techniques to leverage this. We propose 3 such techniques. The Bidirectional Feature Extraction method is stronger but more computationally expensive, and its performance is evaluated in Section \ref{dt_performance}. The Fallback Method leverages the likelihood estimation to enhance model prediction, which is an efficient method that only trains the feature aggregation layer and prediction module, leaving the pre-trained GPT intact. This method is evaluated in Section \ref{PEFT}.  Finally, Test-time Domain Adaptation is a post fine-tuning method designed to improve generalization capabilities of the model, evaluated in Section \ref{generalization}. 

\subsubsection{Bidirectional Feature Extraction}\label{bfe} A key disadvantage of GPT compared to transformer encoder methods is that due to its training paradigm, we need to disable bidirectional attention to prevent information leakage. However, when we want to extract features of each patch and aggregate to sequence level feature, bidirectional information may be helpful. We therefore take inspiration from LLM2Vec  \cite{behnamghader2024llm2veclargelanguagemodels}, which demonstrates that in language domain, GPT models can be made bidirectional and surpass the performance of encoder-based approaches in terms of feature extraction. Different from LLM2Vec, which employs a 2-step training process to adapt GPT, we simply modify how the signal modeling loss is computed in the proposed mixed objective fine-tuning framework, while keeping the feature aggregation and prediction module (the lower branch in Figure \ref{f1}) intact. Specifically, after disabling causal attention to allow bidirectional information, we randomly mask out a portion of the input patches and the model is fine-tuned to reconstruct the patches preceding the mask. That is, for a randomly chosen position $i\ge0$, we mask out the patch at position $i+1$ since otherwise the bidirectional attention would leakage the ground truth information, and we compute loss using the model predicted $\mu, b$ at position $i$ and the masked-out patch $i+1$. It is important to note that as before, the model is fine-tuned on both signal modeling and the downstream task objective simultaneously. While this approach leads to higher computational cost (from both bidirectional attention and lower sample efficiency), it does exhibit stronger downstream task performance on various tasks compared to the vanilla approach, thanks to the bidirectional information. We empirically validate its performance in Section \ref{dt_performance}.

\subsubsection{Fallback Method}\label{fallback}
A unique advantage of GPT models is that we naturally obtain a sequence level likelihood. Such information is not easily available for models that aim to perform mask reconstruction like BERT, since we can only compute the likelihood of a subset of patches in a single forward pass. To leverage this, we propose the fallback method. Let $x$ be an input PPG sequence, $L(x)$ be the likelihood of $x$ computed from signal modeling loss, and $P(x)$ be the output of the predictor module, then we define:
$$\hat{y} = L(x)P(x) + \frac{y_{\text{fallback}}}{\max(L(x), \delta)}$$
where $y_{\text{fallback}}$ is a trainable parameter in the same shape as $P(x)$, and $\delta$ is a small positive number preventing numerical issues in the denominator. Specifically, for regression tasks, $y_\text{fallback} \in \R$, and for a $C-$way classification task, $y_\text{fallback} \in \R^C$, which is added to model predicted class logits (this ensures that after softmax, the final output is a valid distribution over $C$ classes). In our experiments, we set $\delta$ to be 0.1 and we found that $\delta$ is rarely invoked. Intuitively, for a high likelihood $L(x)$, the model mainly relies on $P(x)$ since $y_{\text{fallback}}$ gets scaled to a small number. But for a low likelihood, the contribution of $P(x)$ diminishes and $y_{\text{fallback}}$ term grows in magnitude. 

Importantly, this approach relies on the likelihood estimation from a pre-trained model. Fine-tuning the entire model using the framework described in Section \ref{mixed_objective} defeats the purpose, because then the model is fine-tuned to minimize signal modeling loss $-\log L(x)$ as a part of the objective, which is equivalent to maximizing $L(x)$. Indeed, empirically we observe no significant difference in performance regardless of whether this method is applied or not when performing full-parameter fine-tuning. However, the advantage of this approach manifests when we freeze the GPT layers and train the prediction module and feature aggregation layer only. We found that fallback approach is a sound choice when full-parameter fine-tuning is too expensive. We show its strong performance against vanilla interpolation (i.e., freeze GPT) on various downstream tasks in Section \ref{PEFT}. 

\subsubsection{Test-time Domain Adaptation}\label{personalization}

The fact that we need both signal modeling loss and objective loss to properly finetune the model implies that GPT-PPG is sensitive to the PPG data distribution. As we show in the next section, this is not an issue when the training distribution matches the test distribution. However, in heart rate estimation tasks where we perform leave-one-session-out validation, the distribution discrepancy among different subjects emerges. Therefore, we further propose a subject personalization technique to alleviate the issue, where we take a small portion of the PPG signals in test set to finetune the model using signal modeling loss only. Note that there is no data leakage since we do not use the labels in the test set for model fine-tuning. This technique adapts the model to the PPG distribution of the test data and only requires very little extra computation since we only use a small subset of the data. We demonstrate the effectiveness of this post fine-tuning method in Section \ref{generalization}. 

\section{Experiments and Results}

Here we present fine-tuning results on various downstream tasks, which includes heart rate (HR) estimation, atrial fibrillation (AF) detection, blood pressure (BP) estimation, and respiration rate (RR) estimation. In the first subsection, we provide experiment results for each of the the above downstream tasks, showing that our model excels in many of them. For HR estimation, we use leave-one-subject-out evaluation (LoSo). For AF detection, we use a test set given by the original dataset, which is split based on subjects. For BP, we randomly selected subset from the whole dataset as the test set. To prevent distribution leakage, we explicitly made sure that the test and training set does not have subject overlap. For RR, we use 5-fold cross validation (CV) since there are too many subjects to perform LoSo. In each CV split, we also make sure that there are no subject overlap bewteen the training and validation set. We then delve into more efficient fine-tuning methods, revealing how we can use GPT's generated representations out-of-box with likelihood estimations. In the third subsection, we investigate our model's ability to generalize to unseen distributions using HR estimation and AF detection as benchmark. Next, we provide empirical evidence that GPT's inherent generative capability can assist the PPG denoising process. Lastly, we report some of findings on the scaling behavior of GPT-PPG. 

It is important to note that different papers handles evaluation in different ways. We therefore specifically include relevant information on evaluation methodology to provide more context for the reported results. In the tables below, CV refers to cross validation and LoSo refers to leave-one-session-out validation. From the results on those downstream tasks, we demonstrate that our model performs exceptionally when we can assume that training and testing distributions match. But when the distributions differ, the performance suffers. This hints at future research in making GPT-PPG more robust. 

\subsection{Downstream Task Performances}\label{dt_performance}

We first report results on AF detection task in Table \ref{AF detection} using the Stanford dataset  \cite{TorresSotoJ}. Coverage refers to the percentage of the original dataset that is retained in evaluation. This process of elimination is usually done by quality-based filtering. As the table shows, we do not perform any filtering as we use the full dataset. The AF detection dataset comes with a predefined train-validation-test split based on subject ID. To the best of our knowledge, all results mentioned in Table \ref{AF detection} are based on the same test set, although when coverage is less than $100\%$, a subset instead of the full test set is used.

\begin{table}[h]
    \centering
    \begin{tabular}{lll}
         \textbf{Method} & \textbf{F1} & \textbf{Coverage}  \\
         \midrule
         DeepBeat   \cite{TorresSotoJ} & 0.652 &  100\% \\
         DeepBeat   \cite{TorresSotoJ} & 0.646 &  27.67\%\\
         Shen et al.  \cite{ShenY} & 0.684 &  100\% \\
         BayesBeat  \cite{DasS} & 0.671 &  100\% \\
         BayesBeat  \cite{DasS} & 0.754 &  54\%\\
         SiamQuality \cite{Ding2024SiamQuality} & 0.710 & 100\%\\
         GPT-19M & 0.77 &  100\% \\
         GPT-85M & 0.823 &  100\% \\
         GPT-345M & \underline{0.841} &  100\% \\
         GPT-1B & \textbf{0.847} & 100\%
    \end{tabular}
    \caption{AF Detection Results (F1)}
    \label{AF detection}
\end{table}

Next, we report results on heart rate estimation task in Table \ref{hr1}. To the best of our knowledge, except for TAPIR \cite{Huang2020}, none of other methods performs quality-based filtering on the dataset -- i.e. removing samples of poor quality. Although BeliefPPG \cite{bieri2023beliefppg} did run experiments that filter out uncertain outputs, the result reported below comes from unfiltered model prediction. While BeliefPPG \cite{bieri2023beliefppg} does outperform our model, we come close and surpasses the other methods by a large margin. 

\begin{table}[h]
    \centering
    \begin{tabular}{lllll}
        \textbf{Methods} & \textbf{WESAD} \cite{SchmidtP} & \textbf{DaLiA} \cite{ReissA} & \textbf{IEEE} \cite{ZhangZ} & \textbf{Evaluation Method}\\
        \midrule
         BeliefPPG \cite{bieri2023beliefppg} & $\underline{4.28\pm 2.0}$ & $\mathbf{3.57\pm 1.4}$ & $\mathbf{1.75\pm 0.8}$ & LoSo \\
         Deep PPG \cite{ReissA} & $7.47 \pm 3.3$& $7.65 \pm 4.2$ & $4.0\pm 5.4$ & LoSo \\
         TAPIR  \cite{Huang2020} & {$\mathbf{4.2\pm 1.4}$} & $\underline{4.6\pm 1.4}$ & $2.5\pm 1.2$ & LoSo with filtering\\
         SiamQuality \cite{Ding2024SiamQuality} & $5.88 \pm 2.34$ & $6.80 \pm 1.87$ & $4.59 \pm 2.93$ & LoSo \\
         GPT-85M &$5.42 \pm 2.2$ & $5.69 \pm 2.4$& $2.72\pm 1.1$ & LoSo \\
         GPT-345M & $5.32 \pm 2.0$& $5.02 \pm 2.1$& $2.33\pm 0.9$& LoSo \\
         GPT-1B & $4.98 \pm 1.8$& $4.77 \pm 1.9$& $\underline{1.98\pm 0.9}$& LoSo 
    \end{tabular}
    \caption{Heart Rate Estimation Results (MAE measured in beats per minute)
    }
    \label{hr1}
\end{table}

We use BIDMC dataset  \cite{PimentelM} to evaluate the model's performance on respiration rate estimation. Since no fixed test set is provided, we report results from 5-fold cross validation. As we can see from Table \ref{rr}, GPT-PPG significantly outperformed other methods and no filtering on the dataset. This consolidates our claim that GPT-PPG is able to generate robust representations for downstream tasks.

\begin{table}[h]
    \centering
    \begin{tabular}{lll}
         \textbf{Method} & \textbf{MAE} & \textbf{Evaluation Method}  \\
         \midrule
         RRWaveNet \cite{osathitporn2023rrwavenet} & $1.66\pm 1.01$ & LOOCV with filtering\\
         RespWatch \cite{dai2021respwatch} & $1.66\pm 2.80$ & 5-fold CV with filtering\\
         LSTM \cite{kumar2022deep} & 1.51 & Random test set\\
         SiamQuality \cite{Ding2024SiamQuality} & $\mathbf{0.89 \pm 0.01}$ & 5-fold CV\\
         GPT-19M & $2.21\pm 0.69$ & 5-fold CV  \\
         GPT-85M & $1.64\pm 0.56$ & 5-fold CV  \\
         GPT-345M & ${1.33\pm 0.48}$ & 5-fold CV  \\
         GPT-1B & $\underline{0.93 \pm 0.21}$ & 5-fold CV \\
    \end{tabular}
    \caption{Respiration Rate Estimation Results (MAE measured in breaths per minute)}
    \label{rr}
\end{table}

Next, we evaluate our models' performance on blood pressure estimation. For this task, we use PulseDB  \cite{WangW} as the dataset, which is a cleaned dataset based on MIMIC-III \cite{johnson2016mimic} and VitalDB  \cite{lee2022vitaldb}. We compare our performance with other methods that use MIMIC-III. Due to the fact that the filtering schemes do not align, the results reported serves as a reference only. Since the dataset is relatively large (483844 samples), we do not perform 5-fold cross validation. The random test set is kept consistent across all of our  experiments. 

\begin{table}[h]
    \centering
    \begin{tabular}{llll}
         \textbf{Method} & \textbf{SBP} & \textbf{DBP} & \textbf{Evaluation Method}  \\
         \midrule
         Slapničar et al.  \cite{slapnicar2019blood} & 9.43 & 6.88 & LoSo with filtering\\
         Kachuee et al.  \cite{kachuee2015cuff} & 12.38 & \textbf{6.34} & Random test set\\
         SiamQuality \cite{Ding2024SiamQuality} & $8.6\pm 6.93$ & -- & 5-fold CV\\
         GPT-19M & 9.13 & 11.8 &  Random test set\\
         GPT-85M & 8.81 &  7.34 & Random test set\\
         GPT-345M & \underline{8.42} &  6.97 & Random test set\\
         GPT-1B & \textbf{8.12} &  \underline{6.78} & Random test set\\
    \end{tabular}
    \caption{Blood Pressure Estimation Results (MAE measured in mmHg)}
    \label{hr0}
\end{table}

Using the 85M model as the base model, we further evaluate the effectiveness of Bidirectional Feature Extraction method. From Table \ref{bfe_vs_plain}, it is clear that this method performs well on several of the downstream benchmarks. For brevity, we only compare the bidirectional approach to the standard unidirectional approach and drop other baselines. We also include percentage of improvement for reference. The signs of \% Change are adjusted according to the metric so that a positive value means improvement and a negative value means a drop in performance. Remarkably, with this technique, 85M parameter model's performance is on par or better than the 1B version on heart rate estimation datasets. 

\begin{table}[h]
    \centering
    \begin{tabular}{lccc}
         \textbf{Dataset} & \textbf{Unidirectional} & \textbf{Bidirectional} & \textbf{\% Change}\\
         \midrule
         Stanford  \cite{TorresSotoJ} & \textbf{0.823} & 0.819& -0.49\%\\
         WESAD  \cite{SchmidtP} & 5.42 & \textbf{4.97}& 8.3\%\\
         DaLiA  \cite{ReissA} & 5.69 & \textbf{4.12}& 27.59\% \\
         IEEE  \cite{ZhangZ} &  2.72 & \textbf{2.08}& 23.53\%\\
         BIDMC  \cite{PimentelM} &  1.64 & \textbf{1.51}& 7.93\%\\
         PulseDB SBP  \cite{WangW} &  \textbf{8.81} & 8.94 & -1.48\%\\
         PulseDB DBP  \cite{WangW} &  \textbf{7.34} & 7.38& -0.54\%\\
    \end{tabular}
    \caption{Unidirectional v.s. Bidirectional Feature Extraction}
    \label{bfe_vs_plain}
\end{table}

In summary, GPT-PPG performs well on different downstream tasks. In the next subsection, we focus more on evaluating the quality of the patch level features that GPT produces, measured by freezing all the GPT layers. 

\subsection{Parameter-Efficient Fine-tuning}\label{PEFT}

In the above experiments, we finetune all of the parameters in the GPT model, which is highly expensive, especially for larger models such as 345M and 1B. Here we provide experimental results where we fix the GPT layers and only train the feature aggregation and prediction module, using GPT-85M as an example. We also explore the impact of using a likelihood-informed prediction approach, detailed in Section \ref{fallback}. We demonstrate in Table \ref{comp} that fine-tuning the full model achieves better performance (see column Unfreeze GPT), but freezing all the GPT layers still yields promising results, especially with the fallback method. This shows that the representation generated by the GPT-PPG model is useful for downstream task. We include the percentage improvement of applying the fallback method when compared against the vanilla Freeze GPT approach. As before, a positive \% Change value means improvment, while a negative means a decrease in performance. 

\begin{table}[h]
    \centering
    \begin{tabular}{l|c|ccc}
         \textbf{Dataset} & \textbf{Unfreeze GPT} & \textbf{Freeze GPT} & \textbf{+ Fallback} & \textbf{\% Change}\\
         \midrule
         Stanford  \cite{TorresSotoJ} & 0.823 & 0.762&\underline{0.766}&0.52\%\\
         WESAD  \cite{SchmidtP} & 5.42 & 7.21& \underline{6.23}&13.59\%\\
         DaLiA  \cite{ReissA} & 5.69 & 8.97&\underline{7.17}&20.07\%\\
         IEEE  \cite{ZhangZ} &  2.72 & 9.02&\underline{7.86}&12.86\%\\
         BIDMC  \cite{PimentelM} &  1.64 & \underline{2.01}&2.23&-10.95\%\\
         PulseDB SBP  \cite{WangW} &  8.81 & \underline{10.99} &11.23&-2.18\%\\
         PulseDB DBP  \cite{WangW} &  7.34 & 8.52& \underline{8.49}&0.35\%\\
    \end{tabular}
    \caption{Freeze GPT v.s. Unfreeze GPT}
    \label{comp}
\end{table}

\begin{figure}[h]
    \centering
    \includegraphics[width=0.6\linewidth]{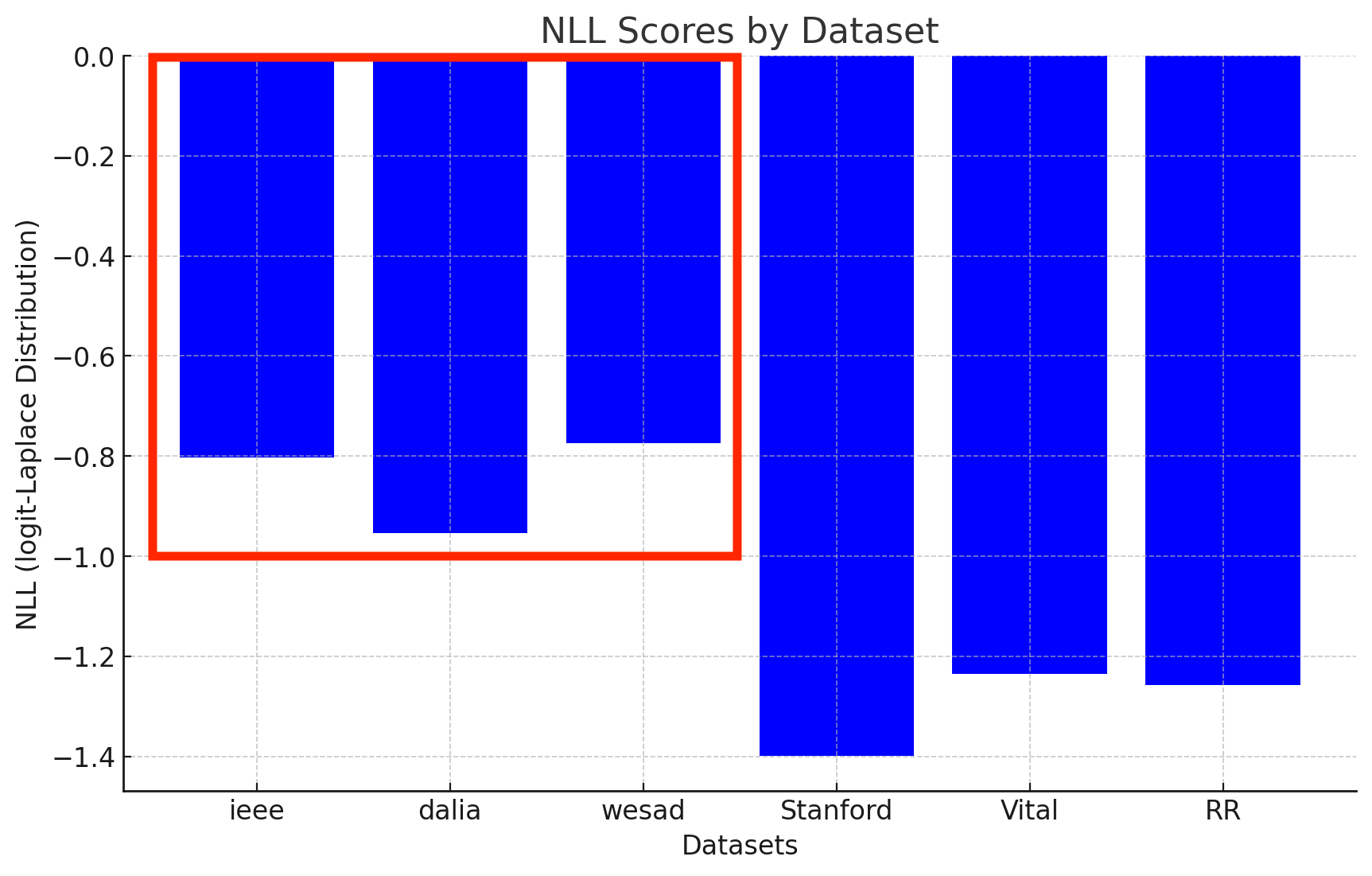}
    \caption{Average Signal Modeling Loss. The datasets that are boxed in red are all heart rate estimation tasks. It is clear that on those datasets, our model is less competitive against the baselines compared to our model's performance on other datasets. }
    \label{loss}
\end{figure}

From Table \ref{comp}, it is clear that HR datasets (WESAD, DaLiA, and IEEE) suffer the most from freezing the GPT. We hypothesize that this is due to their significant distribution differences from the pre-training dataset of the GPT-PPG. To illustrate this difference, we plot the signal modeling loss (Logit-Laplace distribution loss from earlier) of applying the frozen GPT-PPG to the dataset in Figure \ref{loss}. The $y-$axis is the loss value, which is defined to be the negative log likelihood of logit-Laplace distribution. A lower $y-$value means the model is more capable to reconstruct the input data, which implies that the downstream dataset distribution is more aligned with that of the pre-training dataset. Clearly, all the three heart rate estimation datasets (those boxed in red) have higher loss, which implies that the representation provided by the GPT-PPG is less useful. It is also on these datasets that the improvement associated with the fallback method become most significant, while for other datasets the improvement is either marginal or even slightly worse, possibly due to the fact that the likelihood is more uniform and therefore less informative. This demonstrates that for downstream datasets that does not align well with the pre-training data distribution, fallback method could be a robust way to improve the performance, especially when full-parameter fine-tuning is too expensive. 

Note that usually NLL is a positive number because likelihood of a data point given predicted distribution parameters usually does not exceed 1. In fact, this is always the case for discrete distributions since the likelihood in the discrete sense is the same as probability, which cannot be larger than 1. But in our case, the training process encourages the model to learn a very sharp transformed continuous Laplace distribution, which results in a likelihood greater than 1 and therefore a negative NLL. 

\begin{figure}[htbp]
  \centering
  
  \begin{subfigure}[b]{0.48\textwidth}
    \includegraphics[width=\textwidth]{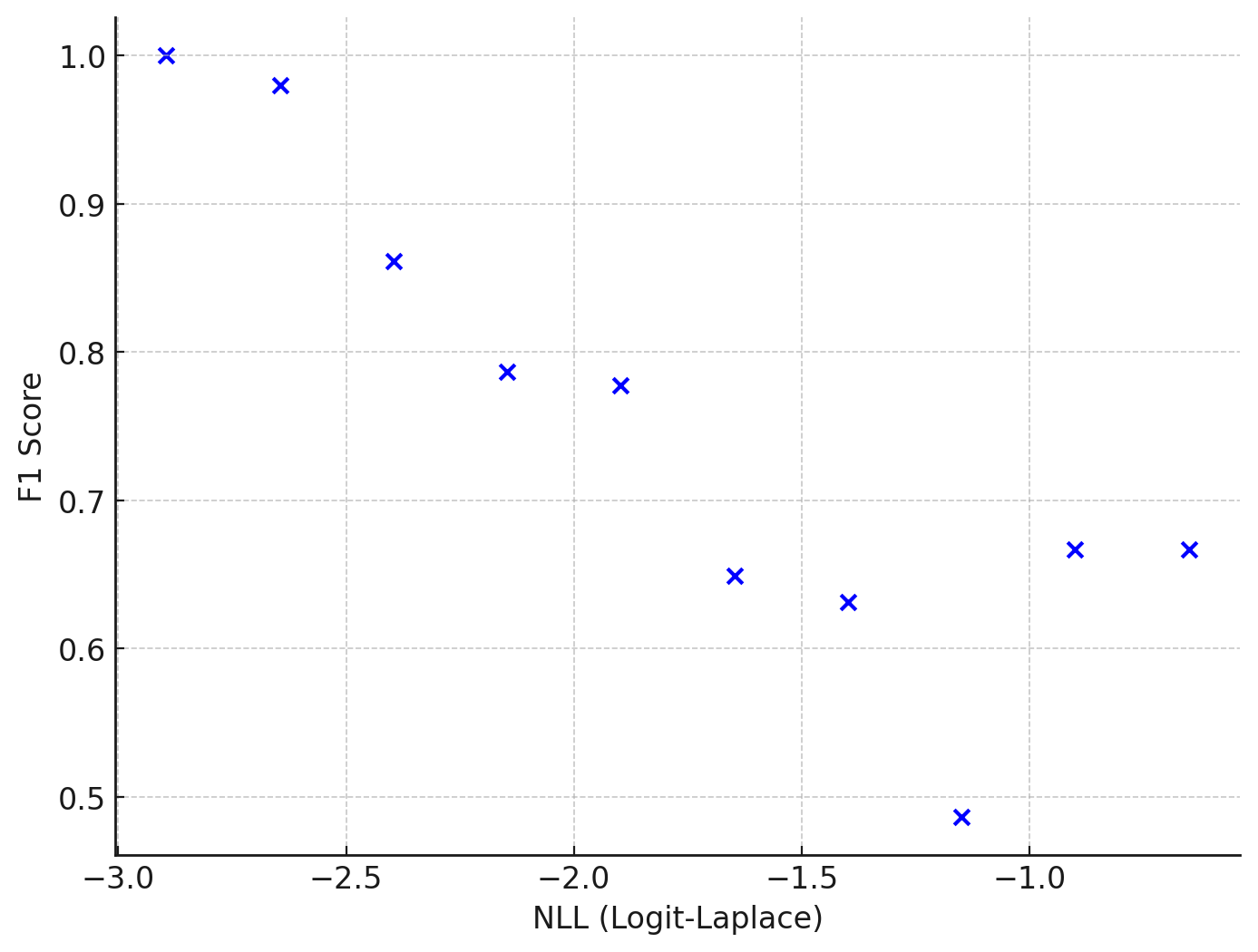}
    \caption{AF Detection}
    \label{fig:sub1}
  \end{subfigure}
  \hfill
  \begin{subfigure}[b]{0.48\textwidth}
    \includegraphics[width=\textwidth]{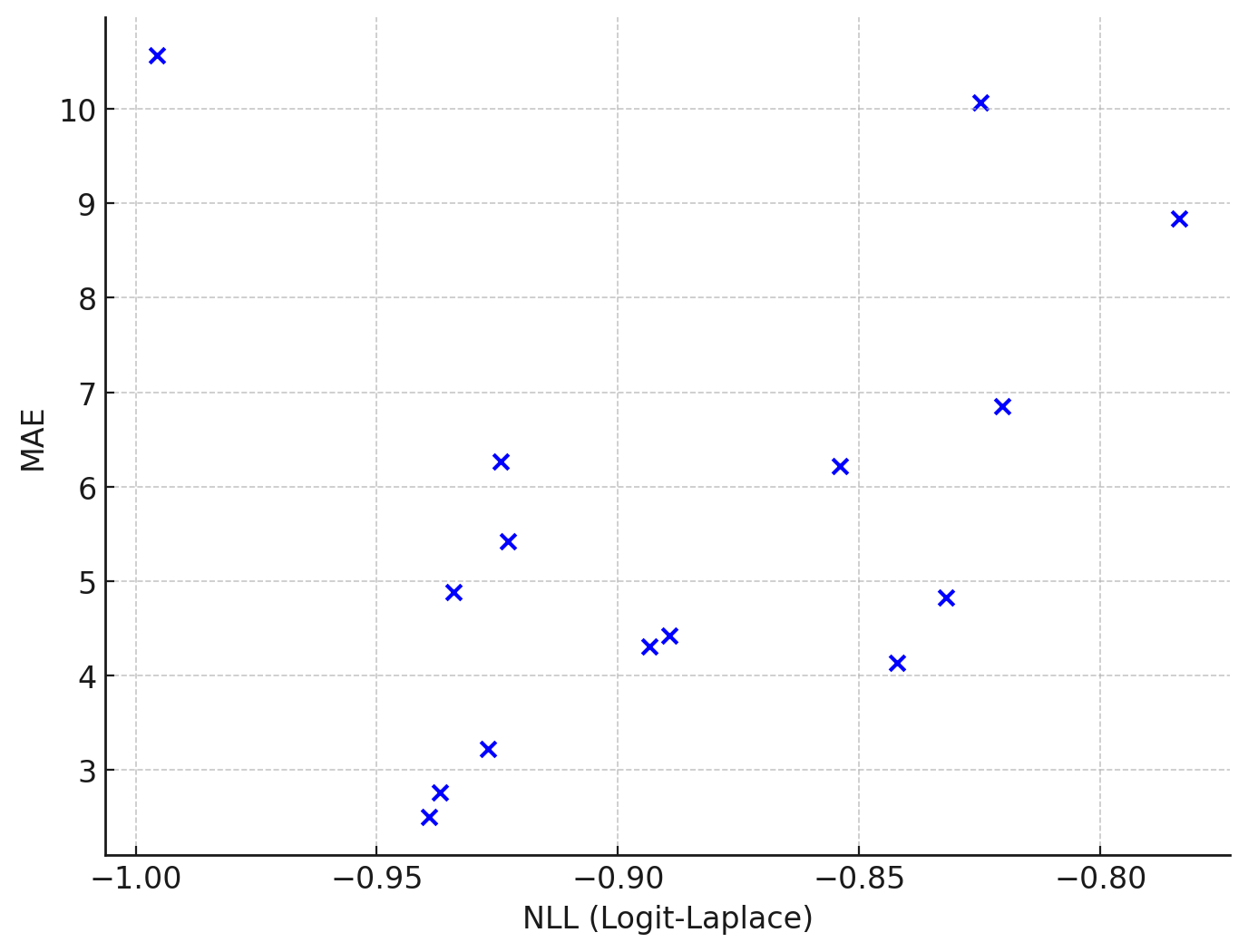}
    \caption{HR Estimation}
    \label{fig:sub2}
  \end{subfigure}

  \caption{NLL v.s. Performance on AF Detection and HR Estimation. Performance of AF detection is measured in F1 score (higher the better) and that of HR estimation is measured in MAE (lower the better). In both datasets, we observe that the performance generally gets worse as NLL increases. }
  \label{AF and HR}
\end{figure}

In addition, we plot the relationship between downstream task performance against NLL on Stanford (AF detection) and DaLiA (heart rate estimation) dataset in Figure \ref{AF and HR}. This further reinforces the observation that the model's capability in reconstructing the sequence (measured by NLL) is highly correlated with its performance. In particular, lower NLL losses correspond to better downstream task performance. 

\subsection{Generalization Studies}\label{generalization}
\subsubsection{Personalization}
On heart rate estimation datasets, while GPT-1B does achieve results comparable to BeliefPPG and TAPIR, the smaller models achieved significantly worse performances. We hypothesize that if GPT-PPG is further adapted to the PPG distribution of the tested subject, we would achieve better performance. We therefore used the test-time domain adaptation technique proposed in Section \ref{personalization} and experimented with GPT-85M. Since the test set is obtained from a single subject, we also call this technique personalization. We report results that use $5\%$ and $10\%$ of the test subject PPG signal in Table \ref{hr2}. 

\begin{table}[h]
    \centering
    \begin{tabular}{lllll}
        \textbf{Methods} & \textbf{WESAD} \cite{SchmidtP} & \textbf{DaLiA} \cite{ReissA} & \textbf{IEEE} \cite{ZhangZ} & \textbf{Evaluation Method}\\
        \midrule
         BeliefPPG \cite{bieri2023beliefppg} & $4.28\pm 2.0$ & $3.57\pm 1.4$ & $1.75\pm 0.8$ & LoSo \\
         GPT-85M &$5.42 \pm 2.2$ & $5.69 \pm 2.4$& $2.72\pm 1.1$ & LoSo \\
         GPT-85M & $4.97\pm 1.9$ & $5.14 \pm 2.1$ & $2.21 \pm 0.9$ & 5\% Personalization\\
         GPT-85M & $4.85\pm 1.8$ & $5.01 \pm 2.0$ & $2.11 \pm 0.9$ & 10\% Personalization\\
         GPT-1B & $4.98 \pm 1.8$& $4.77 \pm 1.9$& $\underline{1.98\pm 0.9}$& LoSo 
    \end{tabular}
    \caption{Heart Rate Estimation Results (MAE measured in beats per minute)
    }
    \label{hr2}
\end{table}

It is clear that personalization improves test-set performance. Note that this personalization technique does not require labels of the test set. That is, we only use the PPG signals from the test subject to perform fine-tuning. This allows the method to become practical in real world scenarios where we have PPG data but lack labeled data. This result demonstrates that GPT is sensitive to the input distribution and that this issue can be alleviated by adapting GPT to the downstream signal distribution. 

\subsubsection{Cross-domain Generalizations}
Next, we present results that measures GPT's ability to generalize across different datasets. Because there may be a significant variation in distributions, this task is challenging. Our results shows that GPT has limited ability in this regard and we plan to pursue this further in future studies. 

In Table \ref{ood results}, we present the performance of the GPT-PPG model on the Stanford atrial fibrillation detection dataset, when fine-tuned on different datasets. The UCSF alarm data consists of over 8 million 30-second PPG segments labeled with AF events generated by automated alarms. The fine-tuning process was conducted under two different settings: one where the GPT layers are unfrozen, allowing both the GPT layers and the classification head to be fine-tuned together, and another where the GPT layers are frozen, limiting fine-tuning to only the classification head. We compare our model against CMC  \cite{CMCPaper}, which is another model trained on UCSF and evaluated on Stanford. From Table \ref{ood results}, it is evident that GPT-PPG performs significantly better when fine-tuned and tested on the same dataset compared to when fine-tuned and tested on different datasets. This indicates that GPT-PPG has limited OOD generalization capabilities.

\begin{table}[h]
    \centering
    \begin{tabular}{llllll}
        \textbf{Fine-tuned Dataset} & \textbf{Model Setting} & \textbf{ACC} & \textbf{F1} & \textbf{AUROC} & \textbf{AUCPR} \\
        \midrule
        UCSF & Unfreeze GPT & 0.480 & 0.313 & 0.477 & 0.240 \\
        UCSF & Freeze GPT & 0.644 & 0.291 & 0.522 & 0.266 \\
        UCSF & CMC \cite{CMCPaper} & - & - & - & 0.735 \\
        Stanford & Unfreeze GPT & \textbf{0.9} & \textbf{0.823} & \textbf{0.936} & \textbf{0.905} \\
    \end{tabular}
    \caption{Performance of GPT-PPG on Stanford AF Detection when fine-tuned on different datasets. The last row represents the scenario where the model's training and test set come from the same source (Stanford).}
    \label{ood results}
\end{table}

\subsection{Empirical Results in Generative Tasks}

\begin{figure}[htbp]
  \centering
  
  \begin{subfigure}[b]{0.32\textwidth}
    \includegraphics[width=\textwidth]{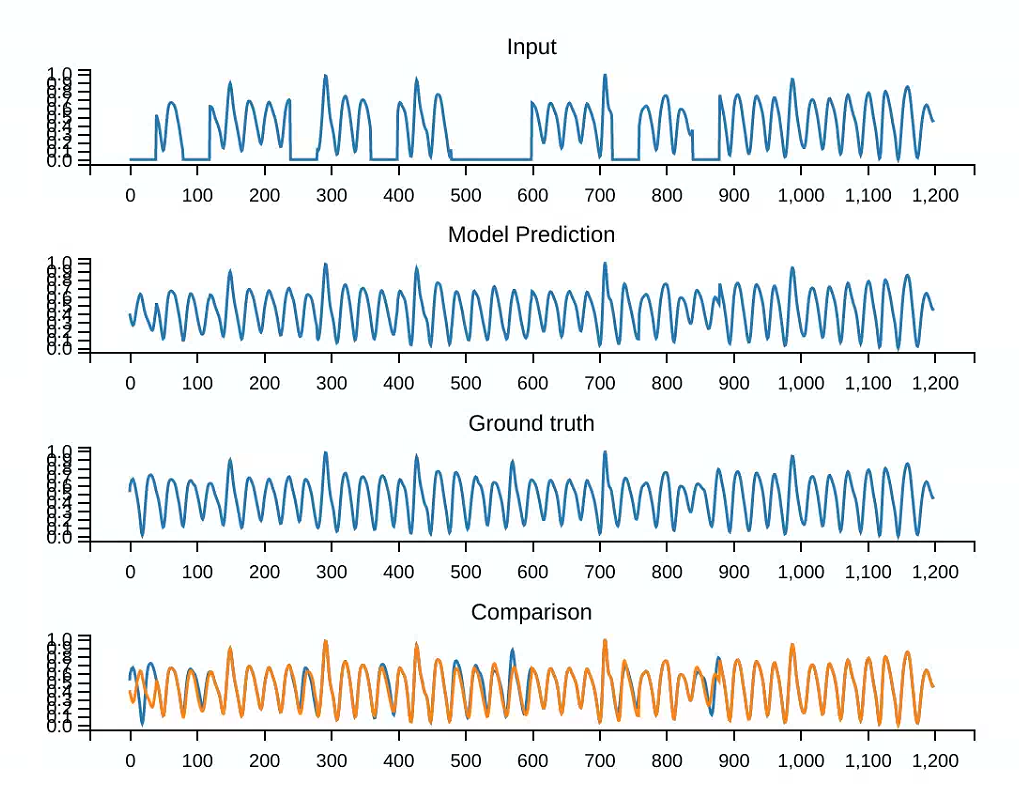}
    \caption{30\% Mask}
    \label{fig:sub1}
  \end{subfigure}
  \hfill
  \begin{subfigure}[b]{0.32\textwidth}
    \includegraphics[width=\textwidth]{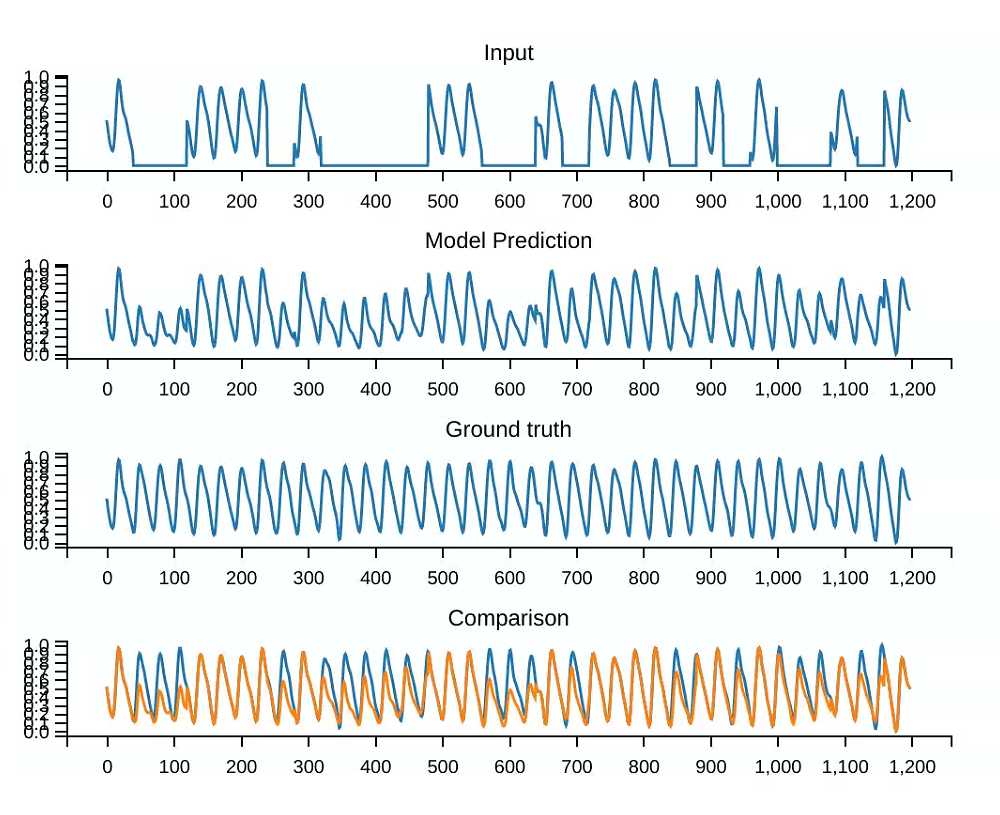}
    \caption{50\% Mask}
    \label{fig:sub2}
  \end{subfigure}
  \hfill
  \begin{subfigure}[b]{0.32\textwidth}
    \includegraphics[width=\textwidth]{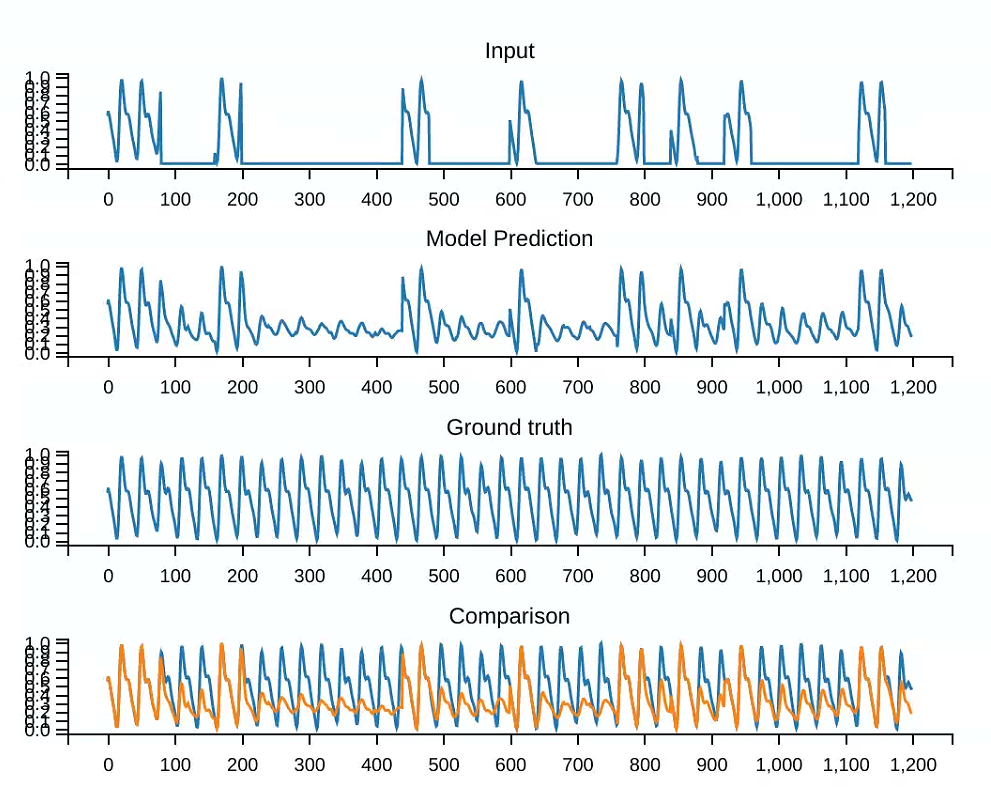}
    \caption{70\% Mask}
    \label{fig:sub3}
  \end{subfigure}

  \caption{Noisy Signal Reconstruction Demonstrations}
  \label{recons}
\end{figure}

Here, we provide some empirical evidence demonstrating the performance of GPT-PPG in performing signal denoising. For each sample, we randomly mask out a portion of the patches by setting them to 0, and we use GPT-PPG to reconstruct those missing patches. This process is sequential -- that is, the model predict patches from left to right, in line with our model pre-training paradigm. In Figure \ref{recons}, we take 3 example signals from the Stanford dataset, which is out of the training distribution, and apply different masking ratios. The model is able to give faithful reconstructions when even 50\% of the original signal is masked. 

\begin{figure}
    \centering
    \includegraphics[width=0.5\linewidth]{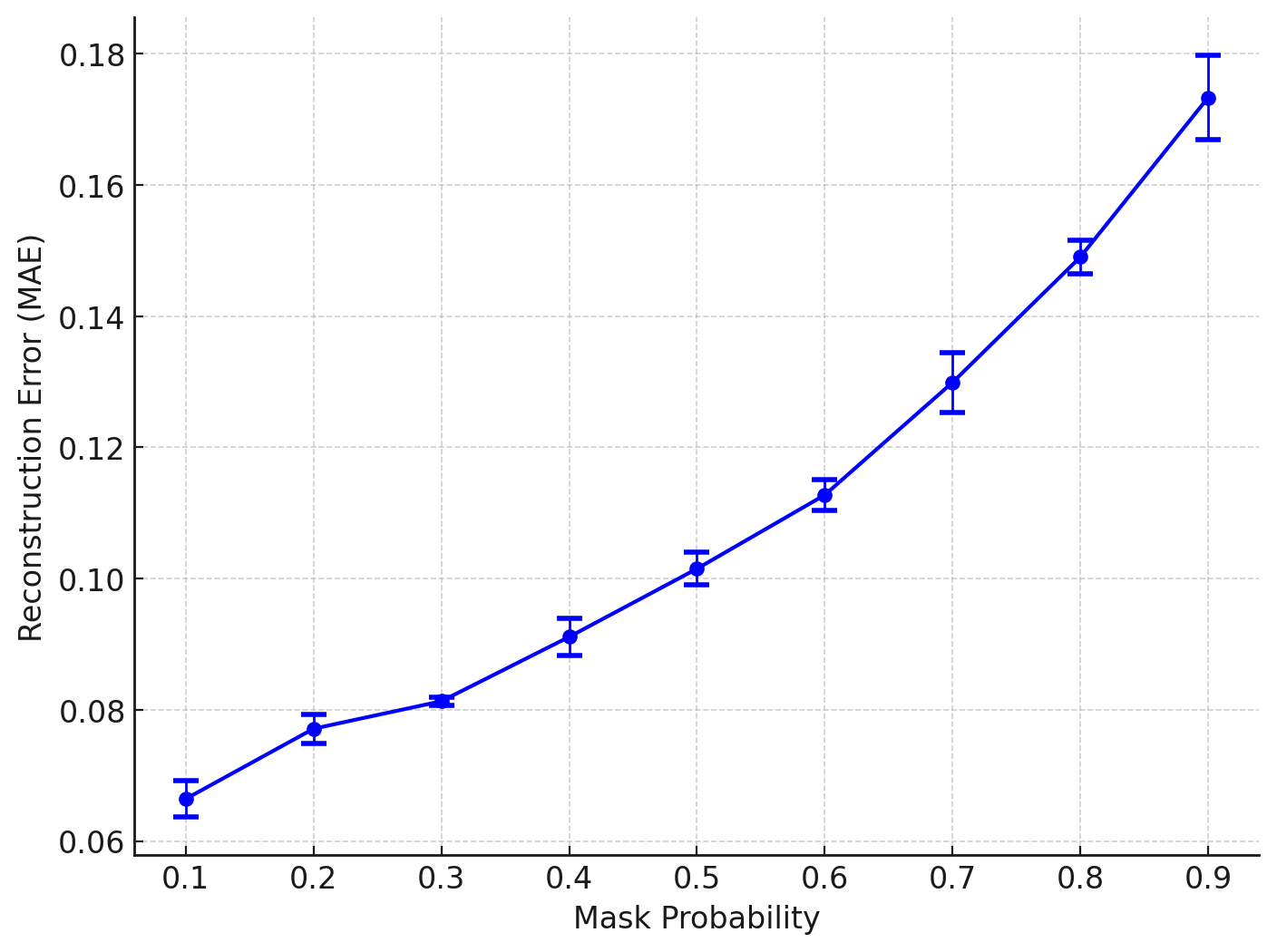}
    \caption{Reconstruction Error v.s. Mask Probability}
    \label{recons2}
\end{figure}

To quantitatively evaluate the model's performance in reconstruction, we plot the model's reconstruction quality with respect to the change in mask ratio (from 0.1 to 0.9) in Figure \ref{recons2}. The results are obtained on a random subset of Stanford dataset, which is out of the pre-training distribution. To provide a more intuitive understanding, the $y$-axis is measured in MAE instead of negative log likelihood. It is clear that the reconstruction MAE increases sharply as more of the original signal is masked out. However, for mask ratios under 40\%, the reconstruction performance is very reasonable as they are all under 0.1 in MAE. It is also important to note that the model being tested here has not been fine-tuned on the Stanford dataset. 

\subsection{Scaling Analysis}
\begin{figure}[h]
    \centering
    \includegraphics[width=0.8\linewidth]{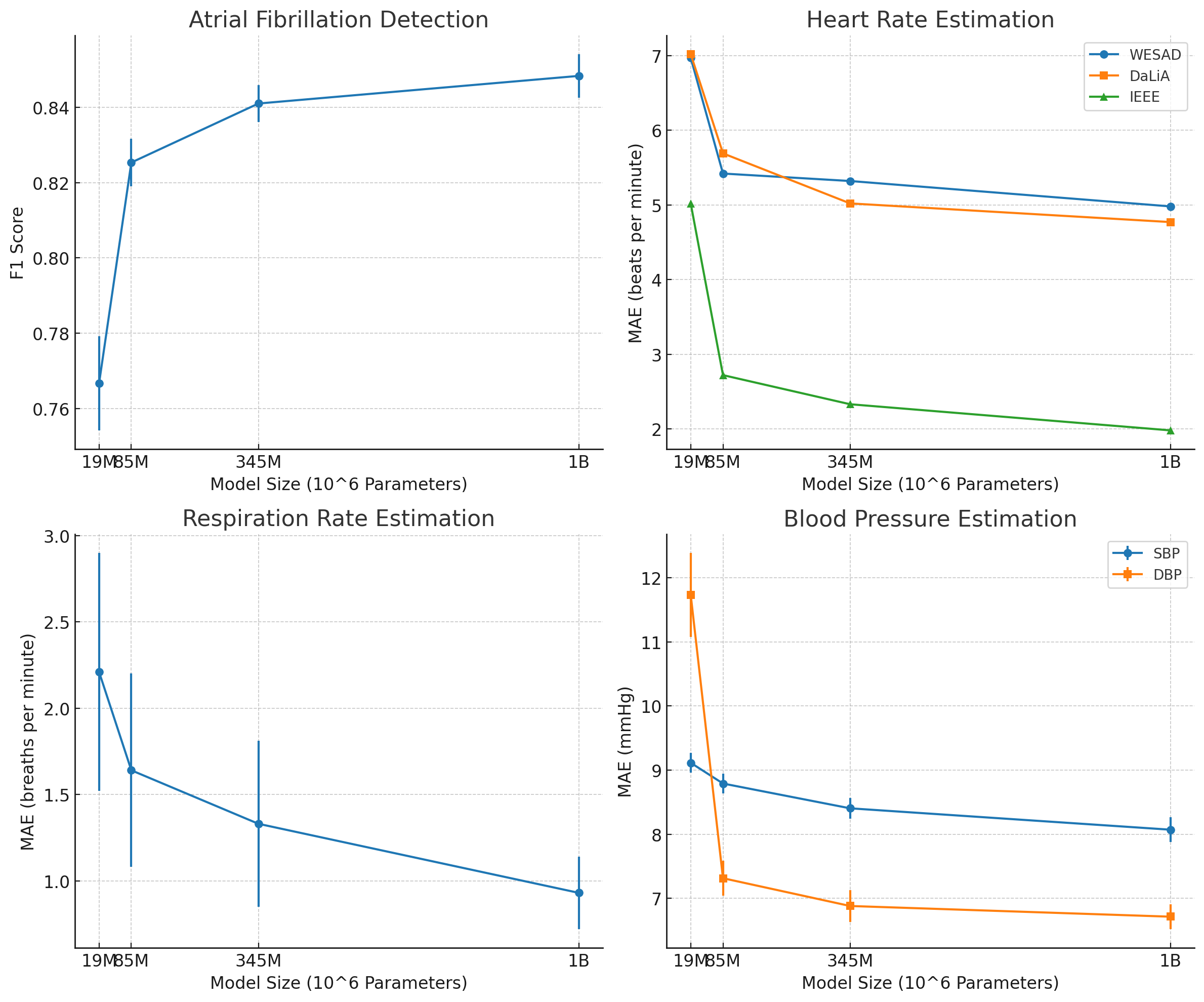}
    \caption{Scaling Curve of GPT-PPG. The error bars are dropped for heart rate estimation tasks for readability (the error bars are on the scale of the mean, like in respiration rate estimation task). We observe much smaller error bars for AF detection and BP estimation, since the error bars in those two datasets are obtained by training and testing the model on the same train/test split multiple times. The variance comes from the inherent randomness of the training process, unlike in RR and HR where the train/test split differ in each trial. The error bar for BP is especially small potentially due to the large size of the dataset. }
    \label{fig1}
\end{figure}

In this work, we introduced GPT-PPG at 4 different sizes: 19M, 85M, 345M, and 1B. Across different tasks, we observed a steady increase of performance as we increase the model size. However, as shown in Figure \ref{fig1}, the improvement of performance is most prominent when going from 19M to 85M model. We believe that 85M strikes the best balance between performance and efficiency, as fine-tuning and running 345M and 1B model poses significant computational challenges. 

\section{Discussion}

This study introduces GPT-PPG, a foundation model for physiological signals, specifically focusing on PPG data, with a generative pre-training objective. Our model demonstrates strong performance across various cardiovascular tasks, including heart rate estimation, respiration rate estimation, blood pressure estimation, and atrial fibrillation detection. It also exhibits promising generative capabilities, such as signal denoising. Our findings offer valuable insights into the strengths and limitations of foundation models for physiological data and provide practical guidelines and methodologies for adapting pre-trained foundation model to specific tasks and resource constraints. While we only used PPG data in this study, this framework can be extended to other modalities. 

\subsection{Strengths of Foundation Models for Physiological Data}

A primary advantage of foundation models in processing physiological signals like PPG lies in their \textit{versatility}. These models can be fine-tuned for a multitude of downstream tasks, achieving impressive results across diverse benchmarks. This versatility is a hallmark of FMs, as evidenced by other successful clinical FM approaches such as  \cite{Ding2024SiamQuality} and  \cite{abbaspourazad2024largescaletrainingfoundationmodels}. In the domain of physiological data, where labeled datasets are often scarce and limited in size, FMs offer a compelling solution. They leverage pre-training on extensive unlabeled datasets to generate rich and informative representations, effectively mitigating the data scarcity issue. Furthermore, beyond feature extraction, GPT-PPG demonstrates utility in signal denoising and personalization, suggesting potential for novel applications akin to the diverse prompt engineering and agent-building methods seen in natural language processing.

\subsection{Adapting Foundation Models for Enhanced Performance and Efficiency}

Adapting foundation models for specific downstream tasks requires careful consideration of several factors. Our experiments with GPT-PPG highlight the crucial role of \textit{data distribution alignment} between the pre-training and downstream datasets. Optimal performance is achieved when these distributions are closely aligned. Unlike encoder models trained with masked reconstruction, GPT provides a mathematically principled method to evaluate this alignment through the chain rule of probabilities, offering an exact likelihood for sequences under the learned distribution \( p_\theta(x) \), an approximation of the implicit distribution \( p(x) \) (see Section \ref{loss_fn description}). Interestingly, the size of the downstream dataset did not significantly impact performance in our experiments, suggesting that GPT-based models can excel even with limited data, as exemplified by the strong performance on the relatively small respiration rate estimation dataset.

The substantial size of foundation models like GPT-PPG presents challenges for effective fine-tuning. \textit{Parameter-efficient fine-tuning (PEFT)} methods provide practical solutions for mitigating this issue. While freezing the model backbone and only fine-tuning the prediction head yields reasonable results, our novel fallback approach (Section \ref{fallback}) significantly improves performance with minimal additional parameters. This method leverages the likelihood information from the pre-trained model, which reflects how well the new signal fits into the learned distribution. In regression tasks, this method requires just one extra trainable parameter, making it an attractive option for edge devices. Future work could explore more advanced PEFT techniques such as applying LoRA  \cite{hu2021loralowrankadaptationlarge}.

\textit{Architectural modifications} can also enhance the performance of foundation models. The unidirectional attention mechanism inherent in the GPT architecture can limit its ability to capture global context. To address this, we introduced Bi-Directional Feature Extraction (Section \ref{bfe}), combining masked next-token prediction with target value prediction during fine-tuning. While computationally intensive, this approach yields superior results, particularly on datasets with poorer alignment with the pre-training data distribution. The improvement is likely due to the model's enhanced ability to integrate information from the entire signal, rather than only past context, leading to more robust representations. This makes it suitable for scenarios demanding high precision and abundant computational resources. 

\subsection{Limitations and Future Directions}

Despite the promising results, our study has several limitations. First, the pre-training of GPT-PPG was conducted on a single, albeit large, PPG dataset. Exploring the impact of more diverse pre-training data on model generalization is a crucial next step. Second, the size of foundation models remains a significant limitation for edge deployment, as models with millions or billions of parameters are challenging to deploy efficiently. Future work should explore the effect of model distillation to smaller sizes that can be deployed on edge devices. Third, a direct comparison between GPT-PPG and other architectures, such as encoder-only models, pre-trained on the same dataset was not performed. Such comparisons would offer valuable insights into the relative strengths and weaknesses of different architectures for physiological signal processing. Fourth, our downstream tasks were limited to cardiovascular applications. Expanding the benchmark to include a wider range of physiological tasks would provide a more comprehensive evaluation of GPT-PPG's capabilities and limitations. Finally, the use of patch-based processing in FMs, including GPT-PPG and others like  \cite{goswami2024moment, NieY}, introduces an assumption of independent predictability of data points within a patch (more details in \ref{loss_fn description}). The validity of this assumption may differ across different signals. While in this work we have not observed a direct implication, this should be an important point to consider when working with other signals that are less structured compared to PPG. A potential solution to this problem is patch-free sequence modeling, which is possible with methods that scales better with sequence length, such as Mamba  \cite{gu2024mambalineartimesequencemodeling}. Addressing these limitations will pave the way for more robust, efficient, and generalizable foundation models for physiological data analysis, unlocking the full potential of AI for healthcare and physiological monitoring.

\section{Conclusion and Future Work}

This work introduces GPT-PPG, a foundation model adapted from GPT to better accommodate the characteristics of PPG signals. Through extensive evaluation, we demonstrated its capability across diverse downstream tasks, with promising results in generative modeling and fine-tuning strategies. However, challenges in generalization to unseen distributions highlight the need for further advancements. Future work will focus on expanding task coverage, improving performance on noisier datasets, and enhancing generalization through more diverse pre-training data.

\section{Acknowledgments}
This work is partially supported by NIH grant award R01HL166233.

\newpage

\bibliographystyle{vancouver}
\bibliography{main.bib}

\begin{thebibliography}{10}

\bibitem{DevlinJ}
Devlin J, Chang M, Lee K, Toutanova K.
\newblock BERT: Pre-training of Deep Bidirectional Transformers for Language Understanding.
\newblock arXiv. 2019;1810.04805.

\bibitem{RadfordA}
Radford A, Narasimhan K, Salimans T, Sutskever I.
\newblock Improving language understanding by generative pre-training.
\newblock OpenAI Blog. 2018.

\bibitem{Hu2024}
Hu X.
\newblock Foundation models for physiological data, how to develop them, and what to expect of them.
\newblock Physiological Measurement. 2024;45(2):020301.
\newblock Published on behalf of Institute of Physics and Engineering in Medicine.

\bibitem{zhou2021informer}
Zhou H, Zhang S, Peng J, Zhang S, Li J, Xiong H, et~al.
\newblock Informer: Beyond efficient transformer for long sequence time-series forecasting.
\newblock Proceedings of the AAAI Conference on Artificial Intelligence. 2021;35(12):11106-15.

\bibitem{wu2021autoformer}
Wu H, Xu J, Wang J, Long M.
\newblock Autoformer: Decomposition transformers with auto-correlation for long-term series forecasting.
\newblock In: Advances in Neural Information Processing Systems (NeurIPS); 2021. p. 101-12.

\bibitem{cao2024tempo}
Cao D, Jia F, Arık SÃ, Pfister T, Zheng Y, Ye W, et~al.
\newblock TEMPO: Prompt-based Generative Pre-trained Transformer for Time Series Forecasting.
\newblock Proceedings of the International Conference on Learning Representations (ICLR). 2024.

\bibitem{garza2023timegpt}
Garza A, Mergenthaler-Canseco M.
\newblock TimeGPT-1.
\newblock arXiv preprint arXiv:231003589. 2023.

\bibitem{NieY}
Nie Y, Nguyen N, Sinthong P, Kalagnanam J.
\newblock A Time Series is Worth 64 Words: Long-term Forecasting with Transformers.
\newblock arXiv. 2023;2211.14730.

\bibitem{goswami2024moment}
Goswami M, Szafer K, Choudhry A, Cai Y, Li S, Dubrawski A.
\newblock MOMENT: A Family of Open Time-series Foundation Models.
\newblock In: Proceedings of the 41st International Conference on Machine Learning. PMLR; 2024. p. to appear.
\newblock Available at \url{https://huggingface.co/AutonLab/MOMENT-1-large}.

\bibitem{SmithM}
Smith M, Fleming L, Geach J.
\newblock EarthPT: a time series foundation model for Earth Observation.
\newblock arXiv. 2024;2309.07207.

\bibitem{heartbeit}
Vaid A, Jiang J, Sawant A, Lerakis S, Argulian E, Ahuja Y, et~al.
\newblock A foundational vision transformer improves diagnostic performance for electrocardiograms.
\newblock NPJ Digital Medicine. 2023;6(1):108.

\bibitem{biot}
Yang C, Westover MB, Sun J.
\newblock BIOT: Biosignal Transformer for Cross-data Learning in the Wild.
\newblock In: Thirty-seventh Conference on Neural Information Processing Systems; 2023. .

\bibitem{frozen_ecg}
Li J, Liu C, Cheng S, Arcucci R, Hong S.
\newblock Frozen Language Model Helps ECG Zero-Shot Learning.
\newblock arXiv preprint arXiv:230312311. 2023.

\bibitem{guo2024siamaflearningsharedinformation}
Guo Z, Ding C, Do DH, Shah A, Lee RJ, Hu X, et~al.. SiamAF: Learning Shared Information from ECG and PPG Signals for Robust Atrial Fibrillation Detection; 2024.
\newblock Available from: \url{https://arxiv.org/abs/2310.09203}.

\bibitem{abbaspourazad2024largescaletrainingfoundationmodels}
Abbaspourazad S, Elachqar O, Miller AC, Emrani S, Nallasamy U, Shapiro I. Large-scale Training of Foundation Models for Wearable Biosignals; 2024.
\newblock Available from: \url{https://arxiv.org/abs/2312.05409}.

\bibitem{charlton2023wearable}
Charlton PH, Allen J, Bail{\'o}n R, Baker S, Behar JA, Chen F, et~al.
\newblock The 2023 wearable photoplethysmography roadmap.
\newblock Physiological Measurement. 2023;44(11):111001.

\bibitem{ding2024photoplethysmography}
Ding C, Xiao R, Wang W, Holdsworth E, Hu X.
\newblock Photoplethysmography based atrial fibrillation detection: a continually growing field.
\newblock Physiological Measurement. 2024;45(4):10.1088/1361-6579/ad37ee.
\newblock Published 2024 Apr 17.

\bibitem{IsmailS}
Ismail S, Siddiqi I, Akram S.
\newblock Heart rate estimation in PPG signals using Convolutional-Recurrent Regressor.
\newblock Computers in Biology and Medicine. 2022;145.

\bibitem{jain2023selfsupervisedalgorithmdenoisingphotoplethysmography}
Jain P, Ding C, Rudin C, Hu X. A Self-Supervised Algorithm for Denoising Photoplethysmography Signals for Heart Rate Estimation from Wearables; 2023.
\newblock Available from: \url{https://arxiv.org/abs/2307.05339}.

\bibitem{10385871}
Ribeiro L, Hu X, Oliveira HP, Pereira T.
\newblock AI-based Models to Predict the Heart Rate Using PPG and Accelerometer Signals During Physical Exercise.
\newblock In: 2023 IEEE International Conference on Bioinformatics and Biomedicine (BIBM); 2023. p. 4398-403.

\bibitem{ChowdhuryM}
Chowdhury M, Shuzan N, Chowdhury M, Mahbub Z, Uddin M, Khandakar A, et~al.
\newblock Estimating Blood Pressure from the Photoplethysmogram Signal and Demographic Features Using Machine Learning Techniques.
\newblock Sensors (Basel). 2020;20(11):3127.

\bibitem{CMCPaper}
Ding C, Guo Z, Rudin C, Xiao R, Shah A, Do DH, et~al.
\newblock Learning From Alarms: A Robust Learning Approach for Accurate Photoplethysmography-Based Atrial Fibrillation Detection Using Eight Million Samples Labeled With Imprecise Arrhythmia Alarms.
\newblock IEEE Journal of Biomedical and Health Informatics. 2024;28(5):2650-61.

\bibitem{huang2023logspectralgan}
Huang J, Zhang W, Ding C, Hu X, Rudin C.
\newblock Log-Spectral Matching GAN: PPG-Based Atrial Fibrillation Detection can be Enhanced by GAN-Based Data Augmentation With Integration of Spectral Loss.
\newblock IEEE Journal of Biomedical and Health Informatics. 2023;27(3):1331-41.

\bibitem{yan2024squwasignalqualityaware}
Yan R, Ding C, Xiao R, Fedorov A, Lee RJ, Nahab F, et~al.. SQUWA: Signal Quality Aware DNN Architecture for Enhanced Accuracy in Atrial Fibrillation Detection from Noisy PPG Signals; 2024.
\newblock Available from: \url{https://arxiv.org/abs/2404.15353}.

\bibitem{Chen_2024}
Chen SF, Guo Z, Ding C, Hu X, Rudin C.
\newblock Sparse learned kernels for interpretable and efficient medical time series processing.
\newblock Nature Machine Intelligence. 2024 Sep;6(10):1132–1144.
\newblock Available from: \url{http://dx.doi.org/10.1038/s42256-024-00898-4}.

\bibitem{Ding2024SiamQuality}
Ding C, Guo Z, Chen Z, Lee RJ, Rudin C, Hu X.
\newblock SiamQuality: a ConvNet-based foundation model for photoplethysmography signals.
\newblock Physiological Measurement. 2024;45(8):085004.

\bibitem{pillai2024papageiopenfoundationmodels}
Pillai A, Spathis D, Kawsar F, Malekzadeh M. PaPaGei: Open Foundation Models for Optical Physiological Signals; 2024.
\newblock Available from: \url{https://arxiv.org/abs/2410.20542}.

\bibitem{chen2024gptppg}
Chen Z, Ding C, Modhe N, Lu J, Yang C, Hu X.
\newblock Adapting a Generative Pretrained Transformer Achieves SOTA Performance in Assessing Diverse Physiological Functions Using Only Photoplethysmography Signals: A GPT-PPG Approach.
\newblock In: AAAI 2024 Spring Symposium on Clinical Foundation Models; 2024. .

\bibitem{ZhangB}
Zhang B, Sennrich R.
\newblock Root Mean Square Layer Normalization.
\newblock arXiv. 2019;1910.07467.

\bibitem{SuJ}
Su J, Lu Y, Pan S, Murtadha A, Wen B, Liu Y.
\newblock RoFormer: Enhanced Transformer with Rotary Position Embedding.
\newblock arXiv. 2023;2104.09864.

\bibitem{RameshA}
Ramesh A, Pavlov M, Goh G, Voss S, Radford A, Chen M, et~al.
\newblock Zero-Shot Text-to-Image Generation.
\newblock arXiv. 2021;2102.12092.

\bibitem{behnamghader2024llm2veclargelanguagemodels}
BehnamGhader P, Adlakha V, Mosbach M, Bahdanau D, Chapados N, Reddy S. LLM2Vec: Large Language Models Are Secretly Powerful Text Encoders; 2024.
\newblock Available from: \url{https://arxiv.org/abs/2404.05961}.

\bibitem{TorresSotoJ}
Torres-Soto J, Ashley E.
\newblock Multi-task deep learning for cardiac rhythm detection in wearable devices.
\newblock npj Digital Medicine. 2020;3(1):1-8.

\bibitem{ShenY}
Shen Y, Voisin M, Aliamiri A, Avati A, Hannun A, Ng A.
\newblock Ambulatory atrial fibrillation monitoring using wearable photoplethysmography with deep learning.
\newblock In: Proceedings of the 25th ACM SIGKDD International Conference on Knowledge Discovery \& Data Mining; 2019. p. 1909-16.

\bibitem{DasS}
Das S, Shanto S, Rahman M, Islam S, Rahman A, Masud M, et~al.
\newblock BayesBeat: Reliable Atrial Fibrillation Detection from Noisy Photoplethysmography Data.
\newblock Proceedings of the ACM on Interactive, Mobile, Wearable and Ubiquitous Technologies. 2020.

\bibitem{Huang2020}
Huang N, Selvaraj N.
\newblock Robust ppg-based ambulatory heart rate tracking algorithm.
\newblock In: Annual International Conference of the IEEE Engineering in Medicine \& Biology Society (EMBC). IEEE; 2020. p. 5929-34.

\bibitem{bieri2023beliefppg}
Bieri V, Streli P, Demirel BU, Holz C. BeliefPPG: Uncertainty-aware Heart Rate Estimation from PPG signals via Belief Propagation; 2023.

\bibitem{SchmidtP}
Schmidt P, Reiss A, Duerichen R, Marberger C, Van~Laerhoven K.
\newblock Introducing WESAD, a multimodal dataset for wearable stress and affect detection.
\newblock In: Proceedings of the 20th ACM International Conference on Multimodal Interaction, ICMI '18. Association for Computing Machinery; 2018. .

\bibitem{ReissA}
Reiss A, Indlekofer I, Schmidt P, Van~Laerhoven K.
\newblock Deep PPG: Large-scale heart rate estimation with convolutional neural networks.
\newblock Sensors. 2019;19(14).

\bibitem{ZhangZ}
Zhang Z, Pi Z, Liu B.
\newblock TROIKA: A general framework for heart rate monitoring using wrist-type photoplethysmographic signals during intensive physical exercise.
\newblock IEEE Transactions on Biomedical Engineering. 2015;62(2):522-31.

\bibitem{PimentelM}
Pimentel M, Johnson A, Charlton P, Birrenkott D, Watkinson P, Tarassenko L, et~al.
\newblock Toward a robust estimation of respiratory rate from pulse oximeters.
\newblock IEEE Transactions on Biomedical Engineering. 2016;64(8):1914-23.

\bibitem{osathitporn2023rrwavenet}
Osathitporn P, Sawadwuthikul G, Thuwajit P, Ueafuea K, Mateepithaktham T, Kunaseth N, et~al.. RRWaveNet: A Compact End-to-End Multi-Scale Residual CNN for Robust PPG Respiratory Rate Estimation; 2023.

\bibitem{dai2021respwatch}
Dai R, Lu C, Avidan M, Kannampallil T.
\newblock RespWatch: Robust Measurement of Respiratory Rate on Smartwatches with Photoplethysmography.
\newblock In: Proceedings of the International Conference on Internet-of-Things Design and Implementation (IoTDI '21). Charlottesville, VA, USA: Association for Computing Machinery; 2021. p.~13.

\bibitem{kumar2022deep}
Kumar AK, Majumdar R, Han L, Guo S, Chandra R.
\newblock Deep learning for predicting respiratory rate from biosignals.
\newblock Computers in Biology and Medicine. 2022;141:105338.

\bibitem{WangW}
Wang W, Mohseni P, Kilgore K, Najafizadeh L.
\newblock PulseDB: A large, cleaned dataset based on MIMIC-III and VitalDB for benchmarking cuff-less blood pressure estimation methods.
\newblock Front Digit Health. 2022.

\bibitem{johnson2016mimic}
Johnson A, Pollard T, Mark R. MIMIC-III Clinical Database (version 1.4). PhysioNet; 2016.
\newblock Available from: \url{https://doi.org/10.13026/C2XW26}.

\bibitem{lee2022vitaldb}
Lee H, Jung C. VitalDB, a high-fidelity multi-parameter vital signs database in surgical patients (version 1.0.0). PhysioNet; 2022.
\newblock Available from: \url{https://doi.org/10.13026/czw8-9p62}.

\bibitem{slapnicar2019blood}
Slapničar G, Mlakar N, Luštrek M.
\newblock Blood Pressure Estimation from Photoplethysmogram Using a Spectro-Temporal Deep Neural Network.
\newblock Sensors (Basel). 2019 Aug;19(15):3420.

\bibitem{kachuee2015cuff}
Kachuee M, Kiani MM, Mohammadzade H, Shabany M.
\newblock Cuff-less high-accuracy calibration-free blood pressure estimation using pulse transit time.
\newblock In: Proceedings of the 2015 IEEE International Symposium on Circuits and Systems (ISCAS). Lisbon, Portugal: IEEE; 2015. p. 1006-9.

\bibitem{hu2021loralowrankadaptationlarge}
Hu EJ, Shen Y, Wallis P, Allen-Zhu Z, Li Y, Wang S, et~al.. LoRA: Low-Rank Adaptation of Large Language Models; 2021.
\newblock Available from: \url{https://arxiv.org/abs/2106.09685}.

\bibitem{gu2024mambalineartimesequencemodeling}
Gu A, Dao T. Mamba: Linear-Time Sequence Modeling with Selective State Spaces; 2024.
\newblock Available from: \url{https://arxiv.org/abs/2312.00752}.

\end{thebibliography}
\newpage
\appendix
\section{L1 Loss and NLL of Laplace Distribution}\label{A1}
Here we prove that minimizing the L1 loss between ground truth $y$ and model prediction $\hat{y}$ is equivalent to minimizing negative log likelihood (NLL) of observing $y$ given model prediction $\mu$ and $b$, the location and scale parameter of a Laplace distributed variable. That is, the predictions that minimize NLL also minimizes L1. 

\begin{proof}
    First, by definition of L1 loss, we have
    \begin{equation}\label{l1}
        L_1(x, \hat{x}) = \sum_{i = 1}^N |x_i-\hat{x}_i| = \sum_{i = 1}^N \varepsilon_i
    \end{equation}
    where $N$ is the number of samples and $\varepsilon_i = |x_i-\hat{x}_i|$. The likelihood of observing $x$ given a Laplace variable parameterized by model predicted $\mu, b$ is given by
    \begin{equation}
        \begin{split}
            p(x;\mu, b) &= \prod_{i = 1}^N f(x_i|\mu_i, b_i)\\
            &= \prod_{i = 1}^N \frac{1}{2b_i}\exp\bb{-\frac{|x_i-\mu_i|}{b}}
        \end{split}
    \end{equation}
    Taking NLL gives:
    \begin{equation}\label{nll}
        \begin{split}
            \nll = -\log p(x;\mu, b) &= -\sum_{i = 1}^N \bb{-\log 2b_i - \frac{|x_i-\mu_i|}{b_i}}\\
            &= \sum_{i = 1}^N \bb{\log 2b_i + \frac{|x_i-\mu_i|}{b_i}}
        \end{split}
    \end{equation}
    Now we minimize $\nll_i = -\log p(x_i|\mu_i, b_i)$ with respect to $b_i$. Note that
    $$\frac{\partial \nll_i}{\partial b_i} = \frac{1}{b_i} - \frac{|x_i-\mu_i|}{b_i^2}$$
    By setting the partial derivative to 0, we have 
    \begin{equation}
        \frac{1}{b_i} = \frac{|x_i -\mu_i|}{b_i^2} \Longrightarrow b_i = |x_i - \mu_i|
    \end{equation}
    Substituting the optimal $b_i$ back into \ref{nll} gives:
    \begin{equation}\label{nll2}
        \begin{split}
            \nll &= \sum_{i = 1}^N \bb{\log 2|x_i - \mu_i| + \frac{|x_i-\mu_i|}{|x_i - \mu_i|}}\\
            &= \sum_{i = 1}^N \bb{\log 2|x_i - \mu_i| + 1}\\
        \end{split}
    \end{equation}
    Minimizing \ref{nll2} is equivalent to minimizing $|x_i-\mu_i|$ for every $i$. That is, NLL is minimized when model predicted $\mu_i$ is close to $x_i$, which is exactly the same as \ref{l1}. 
\end{proof}

\section{More on Distribution Loss Function}\label{loss_fn description}
Recall that the model is trained to minimize the negative log of:
\begin{equation}\label{a2_loss_fn}
    f(x; \mu, b) = \frac{1}{2bx(1-x)}\exp\bb{-\frac{|\text{logit}(x)-\mu|}{b}}
\end{equation}
To understand this loss function more intuitively, we can draw a parallel with the GPT models that are trained on text data. In a GPT language model, we predict the distribution of token at position $i$ based on preceding tokens; here, we predict distribution of each data point grouped by patches, based on preceding patches. The reason that this method works is a result of the chain rule of probability. With the assumption that any PPG sequence is sampled from an underlying hidden (and potentially very complex) distribution $p(x)$, we can write:
\begin{equation}\label{factorization}
    p(x) = p(x_1)p(x_2|x_1)\cdots p(x_N|x_1, ..., x_{N-1})
\end{equation}
where each $x_i$ is a patch. Note that this factorization is valid for any joint distributions and is not dependent on the domain. Because we do not know $p(x)$, we use a GPT model $M$ with parameter $\theta$ to approximate it as much as possible by maximizing 
\begin{equation}
    p_\theta(x) = \prod_{i = 1}^N F(x_i; M_\theta(x_{<i}))
\end{equation}
where the patch likelihood $F$ for patch $x_i$ is given by
\begin{equation}\label{patch_likelihood}
    F(x_i; M_\theta(x_{<i})) = \prod_{j = 1}^{\text{patch\_size}}f([x_{i}]_j; M_\theta(x_{<i})_j)
\end{equation}
for $x$ sampled from our training dataset and $f$ as defined in Equation \ref{a2_loss_fn}. We use $x_{<i}$ to denote the patches preceding the $i^\text{th}$ patch. Importantly, although the factorization \ref{factorization} works for any joint distribution, the formulation of $F$ has an underlying assumption that each data point within a patch can be \textit{independently} predicted from preceding patches. The only 2 scenarios in which we can multiply likelihood together to get the joint likelihood is either the joint one can be factorized in some way (Equation \ref{factorization}) or that they are independent. We are invoking the second condition in the formulation of \ref{patch_likelihood}. Indeed, observe that when the model predicts $\mu_j$ and $b_j$ for each data point $j$ within patch $i$, we do not use the ground truth data points contained in patch $i$ as conditioning variable. This is not necessarily a trivial assumption, but empirically it has worked quite well in PPG data. One possible explanation is that PPG is highly periodical, which means that the preceding patches may provide sufficient information to predict patch $i$, even without internal information within patch $i$. For numerical stability and convenience, we minimize the average of negative log of $p_\theta(x)$, which is given by:
$$L_{\text{generative}}(x) = -\frac{1}{N\times \text{patch\_size}}\sum_{i = 1}^N \sum_{j = 1}^\text{patch\_size}\log f([x_{i}]_j; M_\theta(x_{<i})_j)$$
After the model is trained, for any sequence $x$, we can compute the likelihood that $x$ falls into the distribution parameterized by $\theta$, which is an approximation of $p(x)$. 
\end{document}